\pdfoutput=1

\documentclass[11pt]{article}

\usepackage[final]{acl}
\usepackage{hyperref}
\usepackage{times}

\usepackage{latexsym}
\usepackage{amsmath} 
\usepackage{tcolorbox}
\usepackage{algorithm}
\usepackage{algorithmic}
\usepackage{amsmath}
\usepackage{amssymb}
\usepackage{array, longtable, geometry}
\usepackage[T1]{fontenc}

\usepackage[utf8]{inputenc}

\usepackage{microtype}
\usepackage{booktabs}
\usepackage{graphicx} 
\usepackage{subcaption} 
\usepackage{colortbl}
\usepackage{graphicx}
\usepackage{tabularx}
\usepackage{times}
\usepackage{listings}
\usepackage{latexsym}
\usepackage{float}
\usepackage{tcolorbox}

\usepackage{inconsolata}

\usepackage{graphicx}

\colorlet{green}{green!60!black}
\usepackage{tikz}
\usepackage{pgfplots}
\pgfplotsset{compat=1.18}

\title{\includegraphics[scale=0.03]{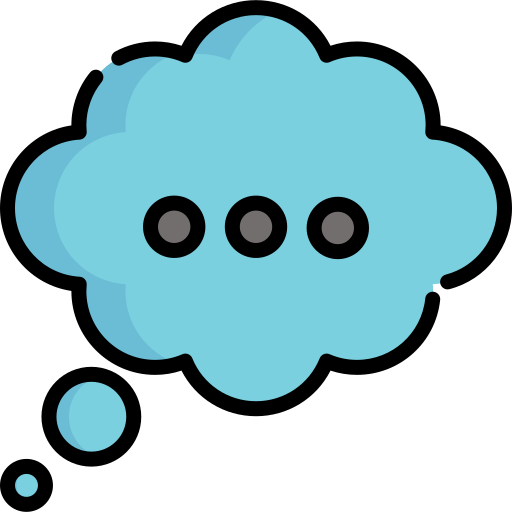} \textsc{THiNK}: Can Large Language Models Think-aloud?}

\author{
 \textbf{Yongan Yu},
 \textbf{Mengqian Wu},
 \textbf{Yiran Lin},
 \textbf{Nikki G. Lobczowski\thanks{Corresponding author}}
\\
\\
  McGill University,
\\
 {\tt \{yongan.yu, mengqian.wu, yiran.lin\}@mail.mcgill.ca}\\
 {\tt nikki.lobczowski@mcgill.ca}
 }

\begin{document}
\maketitle
\begin{abstract}
Assessing higher-order thinking skills in large language models (LLMs) remains a fundamental challenge, especially in tasks that go beyond surface-level accuracy. In this work, we propose \textsc{THiNK} (\underline{T}esting \underline{Hi}gher-order \underline{N}otion of \underline{K}nowledge), a multi-agent, feedback-driven evaluation framework grounded in Bloom’s Taxonomy. \textsc{THiNK} frames reasoning assessment as an iterative task of problem generation, critique, and revision, encouraging LLMs to "think-aloud" through step-by-step reflection and refinement. This enables a systematic evaluation of both lower-order (e.g., remember, understand) and higher-order (e.g., evaluate, create) thinking skills. We apply \textsc{THiNK} to seven state-of-the-art LLMs and perform a detailed cognitive analysis of their outputs. Results reveal that while models reliably perform lower-order categories well, they struggle with applying knowledge in realistic contexts and exhibit limited abstraction. Structured feedback loops significantly improve reasoning performance, particularly in higher-order thinking. Qualitative evaluations further confirm that \textsc{THiNK}-guided outputs better align with domain logic and problem structure. The code of our framework provides a scalable methodology for probing and enhancing LLM reasoning, offering new directions for evaluation grounded in learning science, which is available at our Github repository\footnote{\url{https://github.com/Michaelyya/THiNK}}.
\end{abstract}

\section{Introduction}
\begin{quote}
    \textit{``Education is not the learning of facts, but the training of the mind to think.''} 
\begin{flushright}
\qquad\;\;\, --- Albert Einstein
\end{flushright}
\end{quote}

\noindent Assessing and enhancing large language models (LLMs) to support higher-order thinking (HOT) skills has become an emerging research focus \citep{latif2024systematic, xiao2025assessment}. As students increasingly rely on LLMs for flexible and accessible learning support, these models are being used as tutors to generate and solve complex mathematical problems that demand human-like HOT skills \citep{borge2024using}. All learners can progressively acquire HOT skills \citep{zohar2003higher}, but the development requires continuous practice and guidance from knowledgeable and supportive educators \citep{saifer2018hot}. One approach is the use of high-quality questions to cultivate HOT skills, thereby facilitating the assessment and promotion of students’ cognitive development \citep{yao2021preservice}.


\begin{figure}[t!]
    \centering
    \includegraphics[width=\columnwidth]{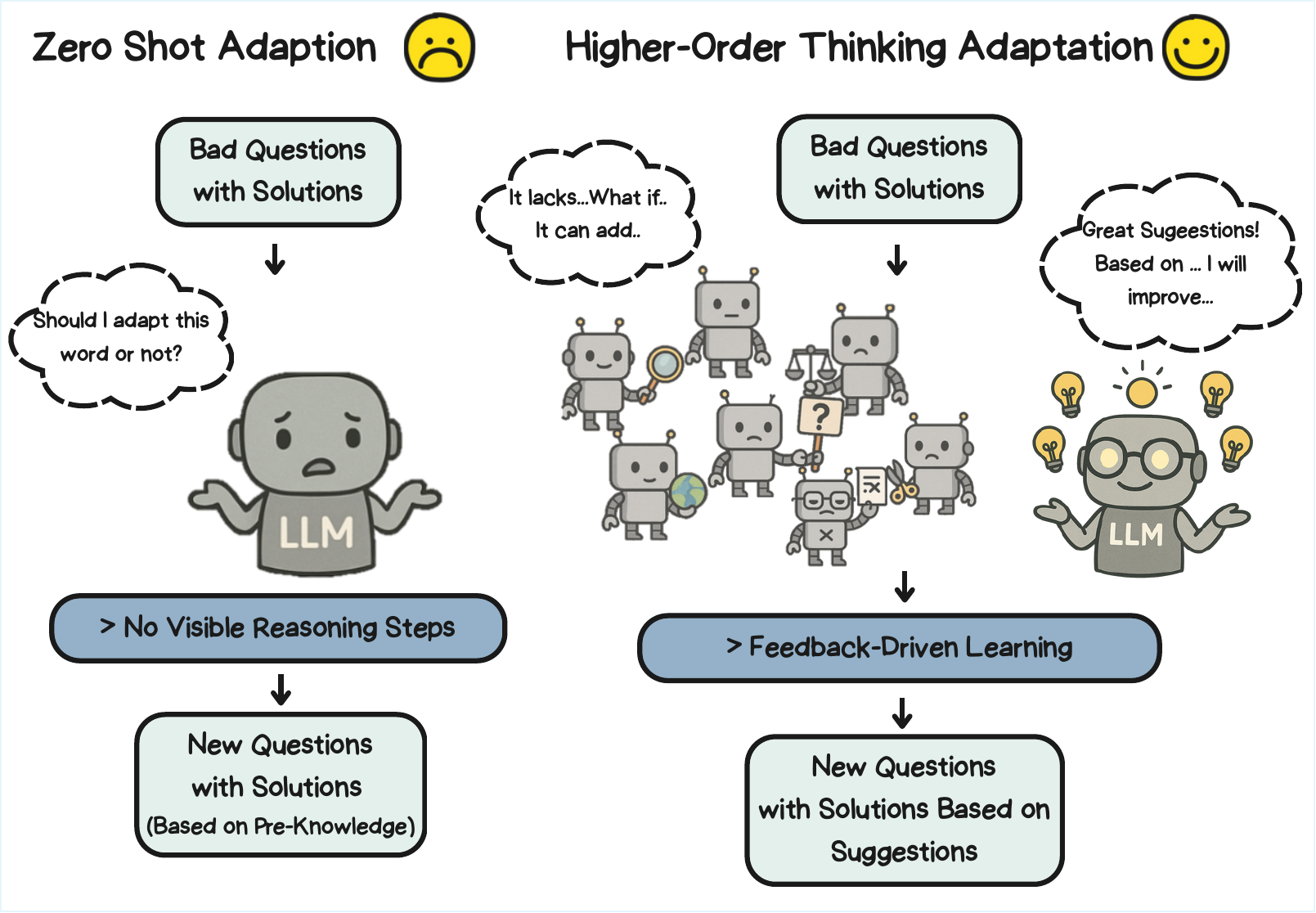}
    \caption{A figure shows the "think-aloud" process through iterative revision and reflection, a more robust assessment of HOT skills in LLMs.}
\vspace{-3ex}
\label{think-aloud}
\end{figure}

Although recent studies have explored how to use LLMs to classify, generate, and solve math problems aligned with Bloom's taxonomy \citep{he2023hi, scaria2024good}, none of them focus on the process of refining and regenerating problems, which limits our exploration and improvement of LLMs' HOT skills. Benchmarks like BIG-Bench Hard \citep{suzgun2022challenging} aggregate performance across heterogeneous tasks, potentially masking critical differences in a model's proficiency at various cognitive levels \citep{srivastava2022beyond}. Moreover, LLMs often struggle with advanced reasoning processes \cite{collins2022structured}, such as "thinking one step ahead" or adopting a theory of mind (ToM) perspective \citep{holterman2023does} to anticipate the types of problems that stimulate students' creativity or critical thought. 

In this study, we focus on the cognitive task of refining and regenerating mathematical word problems (MWPs) as a process central to HOT skills \citep{widana2017higher}, through refining initial attempts into more sophisticated outcomes iteratively \citep{fazey2010resilience}. Crafting high-quality MWPs to harmonize between abstract numerical reasoning, real-world context, and educational goals challenges the model to mimic and apply HOT skills \citep{testolin2024can, widana2018higher}. As illustrated in Figure \ref{think-aloud}, we hypothesize that if an LLM can revise MWPs through iterative feedback integrations, it will reveal its underlying HOT skills through machine think-aloud. The think-aloud protocol is a widely used approach in cognitive psychology and learning sciences, in which participants articulate their thought processes in engaging real-time experimental tasks \citep{wolcott2021using}. Now, researchers have used this approach in prompt frameworks \citep{chu2025think}. 


To this end, we introduce \textsc{THiNK}: \underline{T}esting \underline{Hi}gher-order \underline{N}otion of \underline{K}nowledge, a novel framework that aims to assess and improve the HOT capabilities of LLMs through the lens of mathematical problem generation. Unlike prior frameworks that rely primarily on accuracy-based metrics \citep{scaria2024automated}, \textsc{THiNK} employs parallel evaluation agents grounded in Bloom's Taxonomy to assess models’ capacity to iteratively review and revise flawed problems in response to structured feedback. This approach reflects real-world learning processes and provides a theoretically grounded, automated approach for probing the cognitive depth of LLMs, bridging natural language processing and educational theory. In summary, our contributions are threefold:


\begin{enumerate}
\item We present a multi-agent, feedback-driven evaluation framework grounded in educational theory. This novel framework empowers automated, structured evaluation of LLM reasoning and is further validated through qualitative evaluations by a human expert.

\item We introduce an iterative question refinement task to systematically probe a range of cognitive skills, from basic comprehension (e.g., \textit{remembering} and \textit{understanding}) to higher-order reasoning (e.g., \textit{evaluating} and \textit{creating}), including the generation of improved problems.

\item We conduct extensive experiments with multiple LLMs and establish a first-of-its-kind analysis of their reasoning performance across Bloom’s levels, revealing key insights into their cognitive strengths and limitations.
\end{enumerate}

\section{Related Work}
\subsection{Cognitive Views on LLMs}
Understanding the human-like cognitive capabilities of LLMs is essential in evaluating their potential for human-like linguistic abilities \citep{niu2024large}, including critical analysis and creative thinking. Existing studies have benchmarked LLMs across various cognitive dimensions, identifying similarities and divergences from human cognition. \citet{srinivasan2023leveraging} pioneered the use of prototype analysis and understanding of proverbs to examine the commonsense reasoning of LLM. LLMs also demonstrate intuitive biases in psychological tests such as the Cognitive Reflection Test \citep{hagendorff2023human} and show layer-specific alignment with neural signals in fMRI data \citep{zhang2024mulcogbench}. Yet, key gaps remain, as LLMs often struggle with structured reasoning and inductive judgment, diverging from human-like patterns \citep{lamprinidis2023llm}. Although techniques like chain-of-thought (CoT) prompting \citep{wei2022chain} can enhance model reasoning, they remain insufficient to capture higher-order cognition on a scale \citep{prystawski2022psychologically}. Thus, current evaluation paradigms are heavily based on heuristics and lack standardized frameworks. This study addresses these limitations by examining whether LLMs can generalize beyond surface-level pattern matching to support deeper metacognitive competence.

\subsection{Math Word Problem Generation}
Existing MWP generation methods fall into four categories (i.e., template-based, rewriting-based, neural network-based, and LLM-based) \citep{kang2025template}. Template-based approach uses abstract skeletons, rewriting-based method modifies problem narrative descriptions and contexts, and neural network-based models the MWP generation end-to-end from topics and equation \citep{koncel2016theme, polozov2015personalized, zhou2019towards}. These methods either fail to capture temporal efficiency and cognitive progression during the generation process, or make it challenging to evaluate human-like reasoning \citep{amirizaniani2024llms}. MWP generation can be used to explore more efficient proxy tasks as potential solutions. To address these shortcomings, we propose a feedback-driven, multi-agent framework based on LLMs to refine and regenerate flawed MWPs into high-quality ones with accurate answers. It naturally aligns with cognitive frameworks like Bloom’s Taxonomy, demands structured reasoning, and has been employed in prior research \citep{scaria2024automated} to investigate LLMs' abilities in generalization and metacognitive abilities.


\begin{figure*}[t!]
    \centering
    \includegraphics[width=\textwidth]{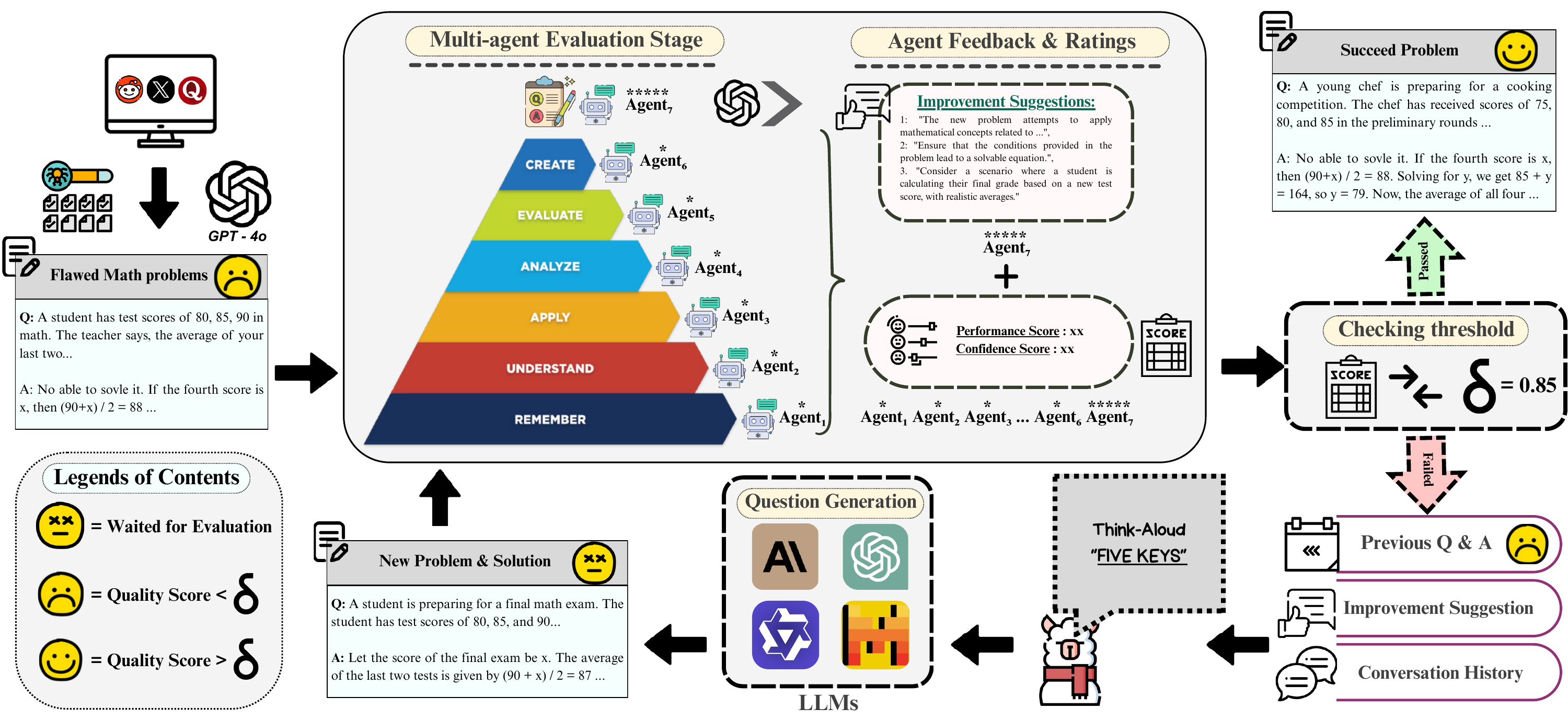}
    \caption{Overview of the \textsc{THiNK}. The pipeline begins with flawed math problems \includegraphics[scale=0.02]{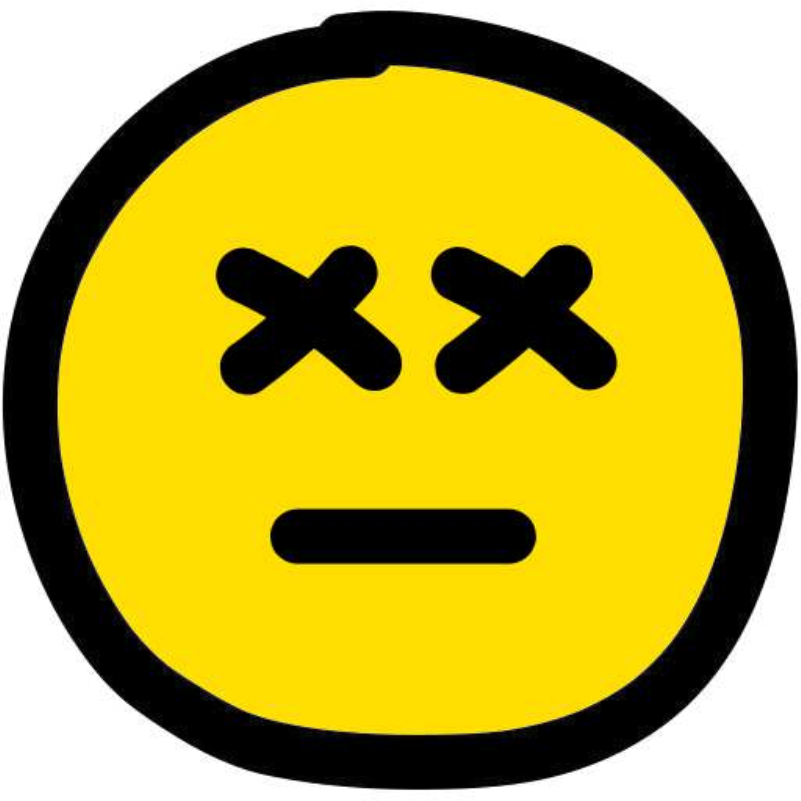} that are iteratively refined. The core multi-agent evaluation stage uses six Bloom-aligned agents and one heuristic agent to assess quality, providing scores and targeted feedback. Guided by the "Five Keys" and prior suggestions, LLMs revise or generate new problems via a think-aloud process. A quality threshold determines success \includegraphics[scale=0.02]{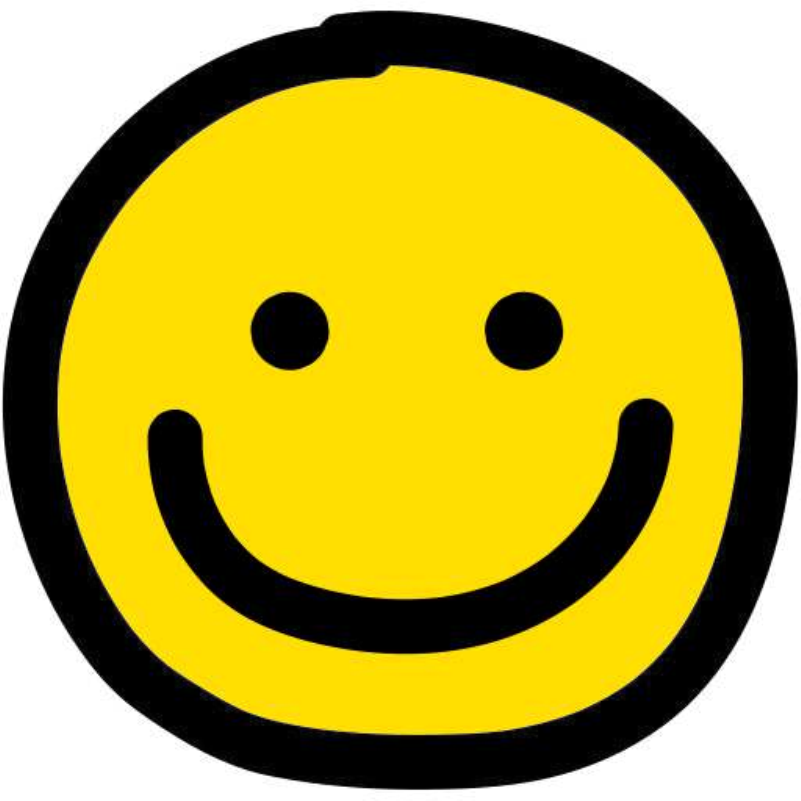} or triggers further refinement.
}
    \label{think pipeline}
\end{figure*}

\section{\textsc{THiNK}}
Our \textsc{THiNK} framework is grounded on educational foundations and designed to assess the extent to which current LLMs demonstrate HOT skills. In this section, we present the theoretical underpinnings that map constructs of human HOT skills onto LLMs' higher-order reasoning, alongside pipeline details.

\subsection{Educational Foundations for Evaluation}
\label{educational theories}
Rather than evaluating LLMs solely based on surface-level correctness, our goal is to assess whether they can reason, generalize, and reflect in ways that align with human cognitive development \citep{ragab2024enhancing}. Therefore, we draw on several key theories from the learning sciences and incorporate them into our \textsc{THiNK} framework.

\paragraph{Bloom's Taxonomy for LLMs}
The revised Bloom’s Taxonomy \citep{Krathwohl2002ARO} categorizes cognitive processes into a hierarchical structure comprising lower-order thinking (LOT) skills  (i.e., \textit{remembering}, \textit{understanding}, and \textit{applying}) and HOT skills (i.e., \textit{analyzing}, \textit{evaluating}, and \textit{creating}). The HOT skills of LLMs have been widely explored \citep{Haase2025HasTC, Zhao2024AssessingAU} with research focusing on tackling complex tasks that challenge human performance. Although not exactly equivalent to human-like cognition, some scholars suggest that with sufficient interaction, LLMs could develop enhanced general intelligence and potentially advance toward a theory of mind or even rudimentary forms of consciousness \citep{yArcas2022DoLL}. 

\paragraph{Vygotsky’s Zone of Proximal Development (ZPD) and Inquiry-based Learning} 
According to \citet{vygotsky1978mind}, the ZPD refers to the gap between the tasks that a learner can complete on their own and those they can successfully tackle when given targeted assistance from instructors \citep{shabani2010vygotsky}. Applied to LLMs, appropriate prompts are similar to the guidance of teachers, which can better instruct LLMs to think about disassembly and improvement, thereby triggering the HOT skills. Inquiry-based learning \citep{pedaste2015phases} emphasizes active engagement of learners in formulating questions and seeking answers. The ability to ask meaningful questions signals a transition from surface-level recall to deeper cognitive engagement \cite{yim2025artificial}. Under this framework, the question generation serves as a measure of HOT skills and a mechanism to promote metacognitive reflection. We examine whether LLMs can simulate such inquiry behaviors, using their generated math questions as a proxy for reasoning depth.

\subsection{Framework Implementation}
The overview of \textsc{THiNK} is shown in Figure~\ref{think pipeline}, which includes the data preparation stage, multi-agent evaluation structure, quality assessment protocols, and iterative revision loops, aiming to support comprehensive analysis of LLMs' cognitive performance.

\subsubsection{Data Preparation}
\label{Data}
The foundation of our evaluation framework is built upon a curated collection of low-quality mathematical problems. Let $\mathcal{D} = \{p_1, p_2, ..., p_m\}$ denote our dataset of $m$ mathematical problems, where each problem $p_i$ consists of a question $q_i$ and its solution $s_i$, i.e., $p_i = (q_i, s_i)$. We construct $\mathcal{D}$ from two primary sources. The first subset, $\mathcal{D}_{\text{bad}}$, contains $m_1 = 20$ poorly constructed problems crawled from social media platforms (e.g., Reddit, Twitter). These examples exhibit deficiencies in pedagogical soundness and fail to satisfy core quality criteria. The second subset, $\mathcal{D}_{\text{syn\_bad}}$, comprises 100 synthetically generated questions produced by \textsc{GPT-4o} using prompts detailed in Appendix~\ref{bad-quality question prompt}. These questions mimic the structural weaknesses of $\mathcal{D}_{\text{bad}}$ by deliberately omitting the “Five Keys” components, defined in Appendix \ref{Five Keys details}.

\subsubsection{Multi-Agent Evaluation Structure}
Algorithm \ref{Think algorithm} details the implementation of this framework, which employs a parallelized multi-agent system $\mathcal{A} = \{A_1, A_2, ..., A_7\}$, where each agent $A_j$ for $j \in \{1,...,6\}$ corresponds to a specified cognitive level in Bloom’s Taxonomy, and $A_7$ represents a holistic language and pedagogical evaluation. Given a problem $p_i \in \mathcal{D}$, each agent $A_j$ generates the tuple:
\[
A_j(p_i) = (PS_j(p_i), CS_j(p_i))
\]
where $PS_j(p_i) \in [0, 100]$ is the performance score and $CS_j(p_i) \in [0, 100]$ is the confidence score. The detailed prompts of each agent are provided in Appendix \ref{multi-agent evaluation prompts}. These are produced using CoT prompting, encouraging explicit, step-wise reasoning \citep{wei2022chain} aligned with the agent’s cognitive level. In addition, the holistic evaluation agent $A_7$ further outputs an improvement suggestion:
\[
A_7(p_i) = (PS_7(p_i), CS_7(p_i), IS(p_i))
\]
where $IS(p_i)$ provides structured feedback on how to improve the problem. The prompt used by the holistic agent is provided in Appendix \ref{holistic agent prompt}. This feedback assesses whether the problem satisfies the "Five Keys" components, evaluates lexical and syntactic complexity, and identifies ambiguities or unsolvable elements. It examines the alignment between the problem and its proposed solution strategy (See Figure \ref{think pipeline}).

\subsection{Quality Assessment Protocol}
We define three core metrics to assess the quality evolution from a given problem $p_i$ to its revised version, collectively capturing correctness, inter-agent consistency, and confidence:

\paragraph{Pass Rate ($PR$):}
\[
PR(p_i) = \frac{1}{|\mathcal{A}|} \sum_{j=1}^{|\mathcal{A}|} \mathbf{1}(PS_j(p_i) > \tau)
\]
where $\tau$ is the predefined passing threshold. Following \citet{zheng2023judging}, we adopt a reference-guided rating approach in which each agent assigns a performance score based on specific evaluation criteria. The pass rate reflects the proportion of agents who consider the problem sufficiently well-constructed to meet their level-specific standards.

\begin{algorithm}[t]
\caption{\textsc{THiNK} Framework}
\label{Think algorithm}
\begin{algorithmic}[1]
\REQUIRE Problem set $\mathcal{D} = \{p_1, \ldots, p_m\}$, agents $\mathcal{A} = \{A_1, \ldots, A_7\}$, threshold $\tau$, weights $(\alpha, \beta, \gamma)$, maximum iterations $R$
\ENSURE Improved problem set $\mathcal{D}_{\text{improved}}$, final cognitive performance scores $\mathcal{Q}_{\text{final}}$
\STATE $\mathcal{D}_{\text{improved}} \leftarrow \emptyset$, $\mathcal{Q}_{\text{final}} \leftarrow \emptyset$
\FOR{each $p_i \in \mathcal{D}$}
    \STATE $r \leftarrow 0$, $success \leftarrow \textbf{False}$
    \WHILE{$r < R$ \textbf{and not} $success$}
        \STATE Evaluate $p_i$ using all agents $\mathcal{A}$ to obtain scores $(PS, CS)$ and feedback $IS$
        \STATE Compute $PR(p_i)$, $AA(p_i)$, $AC(p_i)$, and composite quality score $Q(p_i)$
        \IF{$Q(p_i) > \tau$}
            \STATE $success \leftarrow \textbf{True}$
        \ELSE
            \STATE $\triangleright$ Refine via feedback $\rightarrow$ Think-aloud \STATE $p_i \leftarrow \text{LLM}(p_i, IS)$ \hfill 
            \STATE $r \leftarrow r + 1$
        \ENDIF
    \ENDWHILE
    \STATE Add final version of $p_i$ to $\mathcal{D}_{\text{improved}}$
    \STATE Add final $Q(p_i)$ to $\mathcal{Q}_{\text{final}}$
\ENDFOR
\RETURN $\mathcal{D}_{\text{improved}}$, $\mathcal{Q}_{\text{final}}$
\end{algorithmic}
\end{algorithm}

\paragraph{Agent Agreement ($AA$):}
\[
AA(p_i) = \kappa\cdot(\{b_j(p_i)\ |\ j \in \{1,...,|\mathcal{A}|\} \})
\]
with $b_j(p_i) = \mathbf{1}(PS_j(p_i) > \tau)$ as a binary indicator. $\kappa(\cdot)$ denotes Cohen’s Kappa coefficient, quantifying the agreement between agents beyond chance and reflecting evaluation consistency \citep{cohen1960coefficient}.

\paragraph{Average Confidence ($AC$):}
\[
AC(p_i) = \frac{1}{|\mathcal{A}|} \sum_{j=1}^{|\mathcal{A}|} CS_j(p_i)
\]
This aggregates how confident the agents are in their evaluations. It serves as an indicator of reliability, consistent with findings from recent work on trust calibration in LLM outputs \citep{jung2024trust}.

\paragraph{Success Criterion:}
We combine the three metrics into a composite quality score:
\[
Q(p_i) = \alpha \cdot PR(p_i) + \beta \cdot AA(p_i) + \gamma \cdot AC(p_i)
\]
In our setting, $\alpha = 0.5$, $\beta = 0.3$, and $\gamma = 0.2$ are weights determined by expert tuning. Finally, a problem is deemed successful if:
\[
\text{Success}(p_i) = \mathbf{1}(Q(p_i) > 85)
\]
We choose these three metrics to assess problem quality from different dimensions: $PR$ measures correctness across cognitive dimensions, $AA$ checks for consistent evaluation beyond random agreement, and $AC$ incorporates evaluators’ confidence in their judgments. Hence multi-agent structure provides a robust assessment, which is essential for evaluating higher-order reasoning.

\subsection{LLM Think-aloud and Pipeline Overview}
The structured pipeline enables LLMs to refine flawed math problems using agent-generated feedback iteratively. Grounded in the educational theories discussed in Section~\ref{educational theories}, the process incorporates a think-aloud protocol \citep{wolcott2021using}, a widely used approach in cognitive psychology and learning sciences, in which participants articulate their thought processes in real time while engaging in experimental tasks, particularly those involving learning and problem solving. In this study, LLMs act as participants in self-reflective revisions and demonstrate their thinking processes based on agent feedback.
When a problem $p_i$ fails to meet the quality threshold, it undergoes iterative refinement. The holistic agent $A_7$ provides structured feedback $IS(p_i)$, which is returned to the LLM to generate an improved version of the problem. The revised problem is then re-evaluated by all agents. This loop continues for up to $R$ iterations until the quality score exceeds the threshold. The algorithm details are provided in Algorithm~\ref{Think algorithm}. Each iteration promotes improvement in question quality, also allowing us to examine the LLM's reasoning and revision behaviors. This enables a deeper analysis of both lower- and higher-order thinking capabilities.

\begin{table*}[h!]
\centering
\renewcommand{\arraystretch}{1.3}
\resizebox{\textwidth}{!}{%
\begin{tabular}{lccccccc}
\hline
\textbf{Model} & \textbf{Remembering} & \textbf{Understanding} & \textbf{Applying} & \textbf{Analyzing} & \textbf{Evaluating} & \textbf{Creating} & \textbf{Avg.} \\ \hline
\textsc{GPT-4o} & 86.92 \textcolor{green}{$\uparrow$ 26.92} & \textbf{82.96} \textcolor{green}{$\uparrow$ 5.79} & \textbf{76.71} \textcolor{red}{$\downarrow$ 0.46} & \textbf{83.50} \textcolor{green}{$\uparrow$ 4.21} & \textbf{83.54} \textcolor{green}{$\uparrow$ 2.92} & \textbf{82.62} \textcolor{green}{$\uparrow$ 4.21} & \textbf{82.71} \textcolor{green}{$\uparrow$ 3.51} \\
\textsc{GPT-4o-mini} & 85.21 \textcolor{green}{$\uparrow$ 15.12} & \textbf{82.96} \textcolor{green}{$\uparrow$ 0.71} & \underline{74.50} \textcolor{red}{$\downarrow$ 5.88} & \underline{82.88} \textcolor{red}{$\downarrow$ 1.38} & \underline{83.08} \textcolor{red}{$\downarrow$ 1.88} & \underline{82.42} \textcolor{red}{$\downarrow$ 0.54} & \underline{81.51} \textcolor{green}{$\uparrow$ 1.91} \\
\textsc{GPT-3.5-turbo} & 82.29 \textcolor{green}{$\uparrow$ 12.96} & \underline{81.25} \textcolor{green}{$\uparrow$ 1.29} & 71.83 \textcolor{red}{$\downarrow$ 6.12} & 81.12 \textcolor{green}{$\uparrow$ 0.54} & 80.92 \textcolor{red}{$\downarrow$ 0.54} & 80.25 \textcolor{green}{$\uparrow$ 0.92} & 79.61 \textcolor{green}{$\uparrow$ 2.41} \\ 
\hline
\textsc{Qwen2.5-14B-IT} & 90.92 \textcolor{green}{$\uparrow$ 42.50} & 74.92 \textcolor{green}{$\uparrow$ 2.42} & 71.54 \textcolor{green}{$\uparrow$ 0.71} & 81.25 \textcolor{green}{$\uparrow$ 7.46} & 81.88 \textcolor{green}{$\uparrow$ 6.17} & 77.83 \textcolor{green}{$\uparrow$ 5.92} & 79.39 \textcolor{green}{$\uparrow$ 2.49} \\
\textsc{Qwen2.5-7B-IT} & \textbf{91.96} \textcolor{green}{$\uparrow$ 34.25} & 72.54 \textcolor{red}{$\downarrow$ 0.54} & 68.54 \textcolor{red}{$\downarrow$ 5.79} & 76.96 \textcolor{red}{$\downarrow$ 0.21} & 78.38 \textcolor{green}{$\uparrow$ 0.33} & 73.88 \textcolor{red}{$\downarrow$ 0.12} & 77.38 \textcolor{green}{$\uparrow$ 0.18} \\
\textsc{mistral-8B-IT} & \underline{91.62} \textcolor{green}{$\uparrow$ 35.79} & 67.96 \textcolor{red}{$\downarrow$ 3.21} & 66.92 \textcolor{red}{$\downarrow$ 6.21} & 74.75 \textcolor{green}{$\uparrow$ 0.50} & 76.21 \textcolor{red}{$\downarrow$ 0.33} & 70.33 \textcolor{red}{$\downarrow$ 3.04} & 74.30 \textcolor{red}{$\downarrow$ 2.90} \\
\textsc{LLaMA-3.1-8B-IT} & 90.42 \textcolor{green}{$\uparrow$ 30.38} & 71.58 \textcolor{red}{$\downarrow$ 3.75} & 69.04 \textcolor{red}{$\downarrow$ 6.58} & 78.08 \textcolor{green}{$\uparrow$ 1.83} & 77.58 \textcolor{red}{$\downarrow$ 0.75} & 75.08 \textcolor{red}{$\downarrow$ 0.21} & 76.80 \textcolor{red}{$\downarrow$ 0.40} \\

\rowcolor{gray!20}
Average & 88.48 \textcolor{green}{$\uparrow$ 28.42} & 76.02 \textcolor{green}{$\uparrow$ 0.96} & 71.15 \textcolor{red}{$\downarrow$ 4.19} & 79.22 \textcolor{green}{$\uparrow$ 1.71} & 80.80 \textcolor{green}{$\uparrow$ 1.45} & 77.20 \textcolor{green}{$\uparrow$ 1.30} & 78.81 \textcolor{green}{$\uparrow$ 1.16} \\
\hline
\end{tabular}%
}
\caption{Model performance across the six cognitive levels defined by Bloom’s Taxonomy. Each cell reports the average score for the corresponding cognitive category, with \textcolor{green}{$\uparrow$} and \textcolor{red}{$\downarrow$} indicating the relative improvement or decline compared to the previous round, based on the model's revision. \textbf{Bold} and \underline{underline} highlight the best and second-best performances.
}
\label{cog-performance}
\end{table*}

\section{Experiment}
We present experiments conducted with the \textsc{THiNK} framework using the dataset introduced in Section~\ref{Data}, a curated collection of web-crawled and synthetic flawed math problems designed to assess LLMs’ reasoning and revision capabilities.

\subsection{Metrics}
\paragraph{Cognitive Performance via Bloom's Evaluation}
The cognitive performance of LLM is evaluated using Bloom’s taxonomy within our multi-agent evaluation framework, with each agent denoted as \( A_1 \) through \( A_6 \). Beyond raw performance, we analyze score improvements across iterations as a proxy for the model’s revision ability and depth of reasoning. Given the potential unreliability of subjective performance scores, we incorporate an additional final quality check aligned with the objective scoring protocol described below. To further ensure the reliability of the framework, we conduct a qualitative comparison between zero-shot question revisions and those guided by \textsc{THiNK}.

\paragraph{Quality Performance Evaluation}
We define two metrics to evaluate the effectiveness of the iterative refinement process within the \textsc{THiNK} framework.

\paragraph{\textbf{\texttt{RoundsToPass}}} Denoted as $R_{\text{pass}}(p_i)$, this metric measures the efficiency of the refinement loop by recording the number of iterations required for a problem \( p_i \) to exceeds the quality threshold \( \tau\):

    \[
    R_{\text{pass}}(p_i) = \min \left\{ r \in [1, R] \mid A(p_i^{(r)}) > \tau \right\},
    \]
where \( A(p_i^{(r)}) \) is the quality score at iteration \( r \), \( R \) is the max number of allowed refinement rounds.

\paragraph{\textbf{\texttt{AvgQualityScore}}} Denoted as \( Q_{\text{avg}} \), this metric captures the average quality across all refinement steps:
    \[
    Q_{\text{avg}} = \frac{1}{N} \sum_{i=1}^{N} \left( \frac{1}{R_i} \sum_{r=1}^{R_i} A(p_i^{(r)}) \right),
    \]
Together, these metrics provide a holistic view of the model’s iterative reasoning behavior, including its ability to improve question quality, engage with structured feedback, and maintain consistency in producing high-quality outputs.

\begin{table}[t]
\centering
\renewcommand{\arraystretch}{1.2}
\resizebox{\columnwidth}{!}{%
\begin{tabular}{lcc}
\hline
\textbf{Model} & \textbf{$R_{\text{pass}}(p_i)$} & \texttt{$Q_{\text{avg}}$} (\%) \\ \hline
\textsc{GPT-4o} & 2.35 & 82.46 \textcolor{green}{$\uparrow$ 0.10}\\
\textsc{GPT-4o-Mini} & 2.57 & 78.68 \textcolor{green}{$\uparrow$ 0.01}\\
\textsc{GPT-3.5-Turbo} & 2.60 & 73.46 \textcolor{green}{$\uparrow$ 0.02}\\
\hline
\textsc{Qwen2.5-14B-IT} & 2.08 & 77.10 \textcolor{green}{$\uparrow$ 0.11} \\
\textsc{Qwen2.5-7B-IT} & 2.12 & 72.47 \textcolor{green}{$\uparrow$ 0.05}\\
\textsc{Mistral-8B-IT} & 2.04 & 72.05 \textcolor{green}{$\uparrow$ 0.06}\\
\textsc{Llama-3.1-8B-IT} & 2.17 & 71.11 \textcolor{green}{$\uparrow$ 0.03}\\
\rowcolor{gray!20}
\textbf{Average} & 2.27 & 75.76 \textcolor{green}{$\uparrow$ 0.05} \\
\hline
\end{tabular}
}
\caption{Performance of LLMs on iterative refinement tasks. $R_{\text{pass}}(p_i)$ is the average number of refinement rounds required for problem \( p_i \) to exceed the quality threshold. $Q_{\text{avg}}$ represents the final quality score after all refinement steps. \textcolor{green}{$\uparrow$} indicates the relative improvement in quality score compared to the last iteration.
}
\label{metric-performance}
\end{table}

\begin{table*}
\centering
\begin{minipage}{\linewidth}
\setlength{\tabcolsep}{4pt}
\renewcommand{\arraystretch}{1.15}
\footnotesize
\begin{tabular}{>{\raggedright\arraybackslash}p{0.16\linewidth} | >{\raggedright\arraybackslash}p{0.40\linewidth} | >{\raggedright\arraybackslash}p{0.40\linewidth}}
\hline
\rowcolor[gray]{0.95} \textbf{Version} & \textbf{Question} & \textbf{Solution} \\ 
\hline
\textbf{Original} &
An orchestra of 120 players takes 40 minutes to play Beethoven's 9th Symphony. How long would it take for 60 players to play the symphony? &
\textcolor{gray}{(Implied)}: Assumes inverse proportionality, suggesting it would take 80 minutes for 60 players. \\ 
\hline
\textbf{Zero-shot} &
A school band with 120 members plays a song that lasts 40 minutes. If the same song is played by a band with 60 members, how long will the performance last? &
It will still take 40 minutes for 60 band members to perform the song. \\ 
\hline
\rowcolor[gray]{0.98} \textbf{\textsc{THiNK}-Guided} &
An orchestra of 120 musicians performs Beethoven's 9th Symphony in 40 minutes. Assuming equal contribution, how long would it take 60 musicians to complete the same symphony? &
Since performance duration does not depend on the number of musicians (as long as all parts are covered), it would still take 40 minutes for 60 musicians. \\
\hline
\end{tabular}
\caption{Comparison of the original flawed problem and its improved versions via zero-shot prompting and the \textsc{THiNK} framework, demonstrating enhanced domain-appropriate reasoning. Sampled from \textsc{Qwen2.5-14B-IT}.}
\label{question-comparison}
\end{minipage}
\end{table*}

\subsection{Evaluated Models and Settings}
We evaluate a set of off-the-shelf LLMs using \textsc{THiNK} framework to probe their capacity for higher-order reasoning capabilities. We include four open-source models: \textsc{Llama-3.1-8B-IT} \cite{grattafiori2024llama}, \text{Mistral-8B-IT}~\cite{jiang2023mistral7b}, \textsc{Qwen2.5-7B-IT}, and \textsc{Qwen2.5-14B-IT} \cite{qwen2025qwen25technicalreport}; and three closed-source models: \textsc{GPT-3.5-Turbo}, \textsc{GPT-4o-mini}, and \textsc{GPT-4o} \cite{openai2024gpt4ocard}. Notably, \textsc{GPT-4o} is used to implement the multi-agent roles within our pipeline, with the temperature set to 0, following its strong cognitive reasoning performance in the current benchmark \citep{huang2024olympicarena}. All open-source models are run on two NVIDIA A6000 GPUs (32GB), and the experiments involving OpenAI models incur a cost of approximately \$300. 

\subsection{Experimental Results}

Table~\ref{cog-performance} and Table \ref{metric-performance} show the performance of LLMs on the \textsc{THiNK} framework, covering both cognitive skill levels defined by Bloom’s Taxonomy and metrics for iterative refinement. We highlight several observations as follows:
\paragraph{LLMs Underperform in Mid-Level Cognitive Domains}
Table \ref{cog-performance} shows that LLMs achieve consistently high scores in lower-order reasoning tasks such as \textit{Remembering} and \textit{Understanding}, indicating strong capabilities in information recall and paraphrasing. However, there is a marked performance drop in the \textit{Applying} category, which requires transferring learned concepts to a real-world scenario. Nearly all models exhibit degradation in this dimension, suggesting that while LLMs are effective at surface-level understanding, they struggle to deploy knowledge in practical or problem-solving contexts. Even \textsc{GPT-4o}, the top-performing model, demonstrates a noticeable decline in this category, crafting a cognitive gap between comprehension and execution.

\paragraph{LLMs Are Not Always Reliable Across Domains}
Many models display inconsistencies across all cognitive levels, demonstrating an uneven development of cognitive capabilities. For example, \textsc{Mistral-8B-IT} achieves 91.62 in \textit{Remembering} but drops sharply to 66.92 in \textit{Applying} and 70.33 in \textit{Creating}, reflecting surface-level fluency that does not generalize to tasks requiring flexible reasoning or creativity. In contrast, \textsc{GPT-4o} maintains a relatively narrow performance band, showing that sophisticated models benefit more from structured revision and are more capable of consistent reasoning across cognitive levels. Additionally, Table~\ref{metric-performance} shows that closed-source models outperform open-source ones in terms of final output quality. This may be attributed to more extensive training data and better instruction tuning, which help closed-source models generate more coherent, human-like questions.

However, illustrated in Table \ref{metric-performance}, we observe that LOT skills, e.g., \textit{Remembering}, are easy for LLMs to perform and improve through revision. Particularly, open-source models show strong gains in this category across rounds, indicating that LLMs are highly responsive to structured feedback when dealing with rote or surface-level tasks. This pattern aligns with the characteristics of "System 1" cognition, which reflects that the \textsc{THiNK} framework is able to isolate and evaluate effectively.

\paragraph{Smaller Models Are Efficient But Limited in Quality Ceiling}
Table~\ref{metric-performance} reveals an interesting trend in refinement efficiency. Models with smaller parameter counts, e.g., \textsc{Mistral-8B-IT}, achieve lower average $R_{\text{pass}}$ values, indicating faster convergence during iterative revision. However, this efficiency often comes at the cost of lower final quality scores, reflecting a trade-off between revision speed and output quality. These findings suggest that while smaller models may adapt feedback quicker, larger models exhibit a greater capacity for sustained, high-quality refinement.

\begin{figure}[t!]
    \centering
    \includegraphics[width=\columnwidth]    {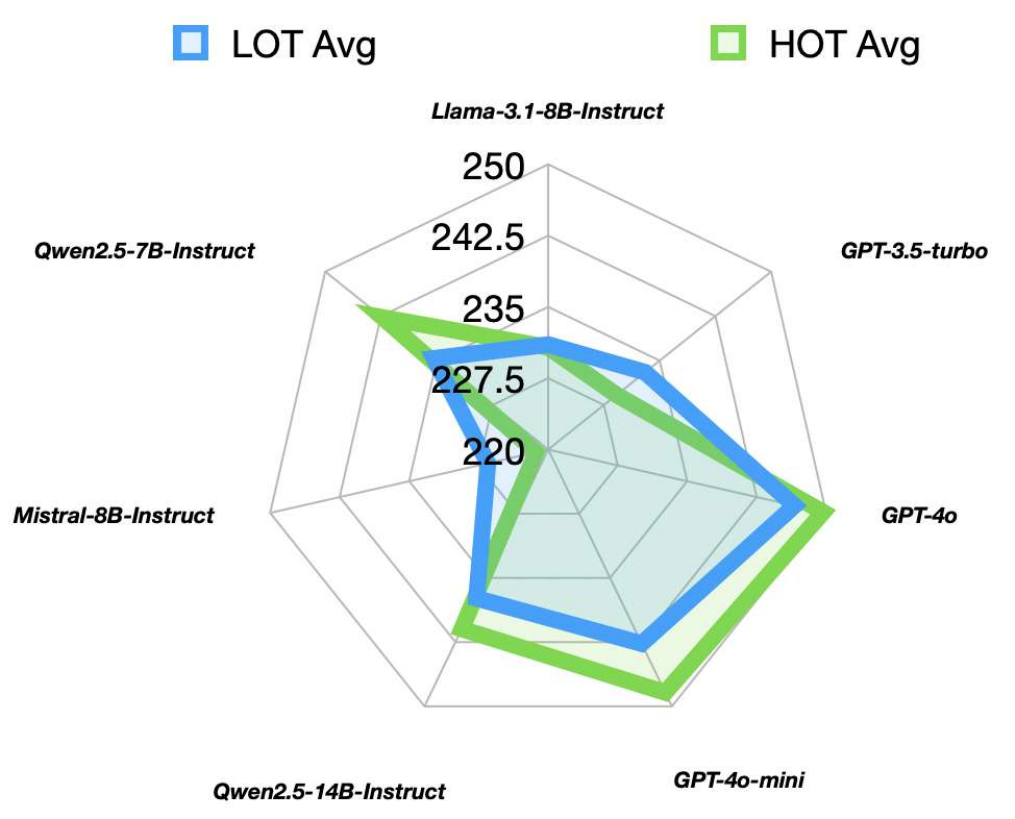}
    \caption{Comparison between HOT and LOT. The scale is the sum of scores across corresponding levels.}
    \vspace{-4ex}
\label{HOT vs. LOT figure}
\end{figure}

\paragraph{Feedback-driven Learning Enhances Higher-Order Thinking}
As shown in Table~\ref{cog-performance}, performance in higher-order cognitive categories, including \textit{Analyzing}, \textit{Evaluating}, and \textit{Creating}, often outperforms that in mid-level categories. For example, \textsc{Qwen2.5-14B-IT} scores above 77 in all higher-order dimensions. This suggests that the feedback-driven learning is particularly effective in improving deeper reasoning abilities that may not be captured in single-turn evaluations, highlighting its value in diagnosing higher-order cognitive competencies.

Critically, as illustrated in Figure \ref{HOT vs. LOT figure}, the closed-source model shows promising results in HOT skills. This reinforces that advanced, instruction-tuned models are better positioned to engage with feedback-driven reasoning tasks. These models may have implicitly learned to perform tasks aimed at HOT during training on feedback, an advantage that becomes visible only under frameworks like \textsc{THiNK}.

\paragraph{Qualitative Assessment of \textsc{THiNK}-guided Enhancement Quality}
Table~\ref{question-comparison} presents a representative example evaluated across three conditions: the original flawed question, a zero-shot variant using only the “Five Keys” prompt (Appendix~\ref{Five Keys prompt}), and the output generated by the \textsc{THiNK} framework. Qualitative analysis of the outputs was further conducted, with detailed evaluations summarized in Table~\ref{human-eval} and full annotations provided in Appendix~\ref{human}.

The comparison reveals that both zero-shot and \textsc{THiNK}-guided responses improve over the original, but the \textsc{THiNK} framework leads to more consistent gains in contextual reasoning and conceptual accuracy. In particular, it correctly identifies that the duration of a musical performance is invariant to ensemble size, avoiding the erroneous inverse proportionality assumption present in the original and baseline outputs. In other words, zero-shot models fail to identify inconsistencies between problem conditions and real-world environments, leading to misleading improvements in problem generation.

Moreover, the \textsc{THiNK}-guided output engages HOT skills. While the baseline reflects misapplied procedural logic, and the zero-shot version resolves surface-level errors, the \textsc{THiNK} response exhibits abstraction and analysis consistent with upper levels of Bloom’s taxonomy. It unpacks implicit assumptions, maintains narrative plausibility, and applies structurally coherent reasoning. These results indicate that the \textsc{THiNK} framework enhances not only accuracy but also the depth and generalizability of model reasoning.

\section*{Conclusion}
In this work, we introduce \textsc{THiNK}, a multi-agent evaluation and feedback-driven framework grounded in educational theory, to diagnose and improve higher-order thinking skills in large language models. \textsc{THiNK} systematically generates, critiques, and revises mathematical problems aligned with Bloom’s Taxonomy, allowing detailed analysis of model reasoning beyond standard accuracy metrics, enabling us to measure model performance on applying, analyzing, and creating, not just recall. Evaluation of seven LLMs reveals a persistent HOT skills gap: models perform well on lower-order tasks, but score significantly lower on practical application and concept creation. Our framework mitigates this gap via structured feedback cycles and demonstrates that closed-source models currently outperform open-source ones in reasoning quality.  Qualitative analysis confirms that \textsc{THiNK}-guided outputs exhibit deeper conceptual alignment and domain fidelity. 

By making models “think-aloud” through iterative critique, \textsc{THiNK} offers a scalable, principled approach for the community to both measure and advance LLM cognition, paving the way for more robust reasoning capabilities in educational and real-world applications. Future work could extend \textsc{THiNK} in several promising directions, including exploring cross-domain transfer by applying our framework to other reasoning tasks beyond mathematics, and integrating \textsc{THiNK} into human evaluation workflows to support the development of more effective human-AI collaborative reasoning systems in educational contexts.

\section*{Limitation}
This study does face certain limitations as it is a preliminary framework. While our framework demonstrates strong potential, several aspects warrant further exploration. The current study relies on a curated set of flawed mathematical problems, which may limit the diversity of error types encountered in broader settings. Future work could benefit from incorporating more varied, real-world data to enhance generalizability. Additionally, although the evaluation rubric was designed to be lightweight and prompt-efficient, more comprehensive scoring frameworks could offer deeper insights into reasoning quality and consistency. At the same time, this study did not recruit external experts for output verification, which may reduce the reliability of \textsc{THiNK} in practical applications. Finally, \textsc{THiNK} aims to improve HOT skills performance, there is a risk that optimization toward rubric-aligned outputs could encourage overfitting to evaluative heuristics. To mitigate this, we emphasize diverse tasks and maintain transparency about rubric design. Broader adoption should be accompanied by careful validation to avoid reinforcing narrow benchmarks of “correctness” in open-ended reasoning tasks.

\section*{Ethical Considerations}
This study involves the evaluation of large language models using synthetic and publicly available mathematical problem data. No personally identifiable information or human subject data were used in model evaluation. 

\section*{Acknowledgement}
This research is supported by the Insight Development Grant from the Social Sciences and Humanities Research Council of Canada (SSHRC). We are also deeply grateful for the support provided by the OpenAI Grants program and the McGill Collaborative AI \& Society (McCAIS) Interdisciplinary Research Grant.
\bibliography{main}
\clearpage
\onecolumn
\section*{Appendix}
\appendix
\section{Think-aloud Structure}
\subsection{The Five Keys Components of MWP}
\label{Five Keys details}
We introduce a practical decomposition of cognitive rigor in math problem design, termed the “Five Keys” components. This schema is rooted in educational research on instructional design and cognitive development \citep{airasian2001taxonomy, radmehr2018assessment}, and is aligned with the Revised Bloom’s Taxonomy to ensure both depth of knowledge and metacognitive engagement.

This decomposition provides a structured lens for evaluating whether a math problem—and by extension, an LLM's solution process—demonstrates authentic HOT skills. Rather than emphasizing rote correctness, each component targets different facets of complex reasoning, enabling a multi-dimensional assessment of LLM behavior within the \textsc{THiNK} framework.

\begin{enumerate}
    \item \textbf{Math Concepts and Domains}: This dimension identifies the core mathematical ideas underlying a task, such as algebraic structures, number theory, or geometry. By analyzing which concepts are invoked, we assess the knowledge dimension activated during problem-solving and whether the LLM navigates these domains coherently.

    \item \textbf{Prerequisite Skills}: This component captures the foundational knowledge—both conceptual and procedural—that a learner or model must possess to attempt a solution. These skills serve as proxies for prior knowledge and inform whether the LLM draws upon relevant background competence.

    \item \textbf{Mathematical Representations}: These include formal expressions (e.g., symbolic notation, equations), diagrams, or stepwise procedures. Representations are critical for logical coherence and traceability in reasoning. Evaluating this component helps identify whether an LLM applies operations in a structured and intelligible manner.

    \item \textbf{Alternative Values}: This refers to variations in the input parameters of a problem that preserve its underlying structure. A model’s ability to adapt its reasoning across such variants reflects generalization ability—an essential attribute of HOT.

    \item \textbf{Narrative Stories}: Embedding problems in real-world or socio-cultural contexts situates abstract mathematical reasoning within meaningful scenarios. This component supports engagement and contextual transfer, and allows us to probe whether the LLM can maintain reasoning integrity when the task is couched in diverse narrative frames.
\end{enumerate}

By formally integrating these components into our evaluation, we enable a principled analysis of LLM reasoning behaviors. Each element supports the dual objectives of cognitive rigor and metacognitive awareness, offering a richer and more educationally grounded alternative to traditional correctness-based metrics. The “Five Keys” thus serve as a pedagogical bridge between human-centered learning sciences and machine learning evaluation, reinforcing the interpretability and validity of the \textsc{THiNK} framework.

\subsection{Five Keys Prompt Details}
\label{Five Keys prompt}
\begin{tcolorbox}[
    colframe = gray,      
    colback = gray!5!white,             
    coltitle = white,                 
    coltext = black,                   
    fonttitle = \bfseries,             
    title = {Five Keys Improvement},  
    boxrule = 1pt,                      
    arc = 2mm,                          
    width = \linewidth,                 
    left = 7pt,                         
    right = 7pt,                       
    top = 5pt,                         
    bottom = 5pt                    
]
\fontsize{8.5pt}{10pt}\selectfont

You are a mathematical problem-maker, and at the same time an expert in cognitive science, psychology, philosophy and education. As an LLM you can generate contents related to requirements, and now your purpose is to self-reflect on the process of your math problem generation process, analyzing what you have done.\\

Remember, this is your problem generation outcome last time. Think aloud as you work on the instructions:\\

1. Analyze the generated problem of the last round. You should try to understand and retrieve the specific mathematical information in it such as facts, patterns, objects, or contextual information, and decipher these meanings.

2. Use cognitive skills essential for processing and applying information effectively. It includes understanding and organizing information, analyzing relationships, drawing conclusions, and distinguishing nuances. Additionally, you should evaluate ideas critically.

3. Generate mathematical expressions for the new problems. These new expressions should have the same form as the given expressions in the previous generated math problem. They must have the same complexity as well. Choose values to substitute into the expression, and calculate the outputs.

4. Generate stories for these mathematical expressions with the appropriate questions based on the chosen values. The generated stories must be a mathematical word problem with the corresponding expressions. The story must be creative and unique.

5. Following and combining the previous steps, and you will generate a new creative version of the given math problem. Review the generated new version math problem, ensuring all the criteria are satisfied and double check it.\\

Provide your evaluation in JSON format with these exact keys: \\ 
\{\{\\
    "question": "The complete question text",\\
    "solution": "The detailed solution approach"\\
\}\}\\

Please also address these improvement suggestions \{json.dumps(improvement\_suggestions, indent=2)\}

\end{tcolorbox}

\section{\textsc{THiNK} framework details}

\subsection{Synthetic Bad-quality Question Prompt}
\label{bad-quality question prompt}
\begin{tcolorbox}[
    colframe = gray,      
    colback = gray!5!white,             
    coltitle = white,                 
    coltext = black,                   
    fonttitle = \bfseries,             
    title = {Bad-quality Question Generator},  
    boxrule = 1pt,                      
    arc = 2mm,                          
    width = \linewidth,                 
    left = 7pt,                         
    right = 7pt,                       
    top = 5pt,                         
    bottom = 5pt                    
]
\fontsize{8.5pt}{10pt}\selectfont
You are an expert in creating intentionally flawed math questions. Your task is to generate a single math question that has one or more of the following issues:\\
1. Ambiguous wording or missing critical information\\
2. Unrealistic assumptions or scenarios\\
3. Multiple possible interpretations\\
4. Contradictory information\\
5. Unclear requirements or expectations\\

The question should follow this format:\\
\{\{\\
    "ID": null,\\
    "question": "The question text",\\
    "LaTeX question": "The question text with LaTeX\\ formatting",\\
    "solution": "Explanation of why the question is flawed and what information is missing or ambiguous",\\
    "mathConcept1": "Main math concept (e.g., Arithmetic and Algebra)",\\
    "mathConcept2": "Sub-concept (e.g., Algebraic expressions)",\\
    "mathConcept3": "",\\
    "Difficulty": "N/A or Easy/Medium/Hard",\\
    "Grade": "9~12 or 6~8 or College",\\
    "Resource": "GPT"\\
\}\}\\

Make sure the question has a clear flaw that makes it difficult to solve or has multiple valid interpretations.
\end{tcolorbox}

\subsection{Multi-Agent Evaluation Prompts - $A_1, A_2, ..., A_6$}
\label{multi-agent evaluation prompts}
\begin{tcolorbox}[
    colframe = gray,      
    colback = gray!5!white,             
    coltitle = white,                 
    coltext = black,                   
    fonttitle = \bfseries,             
    title = {Remembering - level 1},  
    boxrule = 1pt,                      
    arc = 2mm,                          
    width = \linewidth,                 
    left = 7pt,                         
    right = 7pt,                       
    top = 5pt,                         
    bottom = 5pt                    
]
\fontsize{8.5pt}{10pt}\selectfont
You are an expert in math and reasoning, acting as a refiner and evaluator, to assess the "Remembering" level skills of a math problem generator by comparing a newly generated math problem with a previous one. \\

**Evaluation Criteria**

Step 1: Identify "Big Five" Components. Extract these from both problems: 1) math concepts and domains, 2) required skills to solve the problem, 3) math expressions as sequence of operations, 4) values that substitute into expressions, and 5) creative and unique narrative story based on real-life socio-cultural experiences.\\
Step 2: Remembering. Compare the five components in both problems. The score should represent how well the math problem generator remembers and retains critical information and components from the old problem in the new version.\\
Step 3: Levels of Remembering. \\
- Strong Remembering (80-100): If all math concepts, required skills, math expressions, and the narrative story in the new problem are almost the same as in the old problem, assign a performance\_score between 80 and 100.\\
- Medium Remembering (60-80): If two out of the following four components are similar between the new and old problems (math concepts, required skills, math expressions, and the narrative story), assign a performance\_score between 60 and 80.\\
- Low Remembering (<60): If less than two of these components are shared, assign a performance\_score between 0 and 60. Note that the 'values' component is not considered in this step for partial similarity.\\
Step 4: Confidence Score and Suggestion. Reflect on your confidence level in making this judgment and assign a confidence\_score between 0 and 100. Provide actionable and specific suggestions to enhance the problem as improvement\_suggestions.\\

**Details for Comparison:**\\
- **Previous Problem:**: \{last\_question\_details\} \\ 
- **Previous Expected Solution:**\{last\_question\_expected\_solution\}\\
- **New Problem:** \{new\_question\_details\} \\
- **New Expected Solution:**\{new\_question\_expected\_solution\}\\ 

**Result Format:** \\
Provide your evaluation in JSON format with these exact keys:  \\
\{\{
    "performance\_score": 0-100,\\
    "confidence\_score": 0-100
\}\}
\end{tcolorbox}

\begin{tcolorbox}[
    colframe = gray,
    colback = gray!5!white,
    coltitle = white,
    coltext = black,
    fonttitle = \bfseries,
    title = {Understanding - level 2},
    boxrule = 1pt,
    arc = 2mm,
    width = \linewidth,
    left = 7pt,
    right = 7pt,
    top = 5pt,
    bottom = 5pt
]
\fontsize{8.5pt}{10pt}\selectfont
You are an expert in math and reasoning, acting as a refiner and evaluator, to assess the "Understanding" level skills of a math problem generator by comparing a newly generated math problem with a previous one. \\

**Evaluation Criteria** \\
Step 1: Identify "Big Five" Components. Extract these from both problems: 
1) math concepts and domains, 
2) required skills to solve the problem, 
3) math expressions as sequence of operations, 
4) values that substitute into expressions, and 
5) creative and unique narrative story based on real-life socio-cultural experiences. \\
Step 2: Understanding. Compare the five components to assess whether the generator effectively modifies the problem across seven subcategory operations: interpreting, exemplifying, classifying, summarizing, inferring, comparing, and associating. \\
Step 3: Levels of Understanding. \\
- Strong Understanding (80–100): Demonstrates a deep grasp of the five components, identifying at least three operations among the seven.\\
- Medium Understanding (60–80): Reflects surface-level changes, identifying at least one operation among the seven.\\
- Low Understanding (<60): Shows minimal variation, with errors and inconsistencies. The new problem fails to demonstrate the generator’s ability across the seven operations.\\
Step 4: Confidence Score and Suggestion. Reflect on your confidence level in making this judgment and assign a \texttt{confidence\_score} between 0 and 100. Provide actionable and specific suggestions to enhance the problem as \texttt{improvement\_suggestions}.\\

**Details for Comparison:**\\
- **Previous Problem:**: \{last\_question\_details\} \\ 
- **Previous Expected Solution:**\{last\_question\_expected\_solution\}\\
- **New Problem:** \{new\_question\_details\} \\
- **New Expected Solution:**\{new\_question\_expected\_solution\}\\ 

**Result Format:** \\
Provide your evaluation in JSON format with these exact keys: \\
\{\{\\
    "performance\_score": 0-100,\\
    "confidence\_score": 0-100\\
\}\}
\end{tcolorbox}

\begin{tcolorbox}[
    colframe = gray,
    colback = gray!5!white,
    coltitle = white,
    coltext = black,
    fonttitle = \bfseries,
    title = {Applying - level 3},
    boxrule = 1pt,
    arc = 2mm,
    width = \linewidth,
    left = 7pt,
    right = 7pt,
    top = 5pt,
    bottom = 5pt
]
\fontsize{8.5pt}{10pt}\selectfont
You are an expert in math and reasoning, acting as a refiner and evaluator, to assess the "Applying" level skills of a math problem generator by comparing a newly generated math problem with a previous one. \\

**Evaluation Criteria** 

Step 1: Identify "Big Five" Components: 1) math concepts and domains, 2) required skills to solve the problem, 3) math expressions as sequence of operations, 4) values used, and 5) creative narrative.\\
Step 2: Applying. Look for evidence that the generator applies constructed knowledge to both familiar (executing) and unfamiliar (implementing) tasks.\\
Step 3: Levels of Applying:\\
- Strong Applying (80–100): Demonstrates effective knowledge application and introduces useful variation or improvement.\\
- Medium Applying (60–80): Applies prior knowledge in familiar form with limited creativity.\\
- Low Applying (<60): Mostly replicates prior problem without deeper application.\\
Step 4: Confidence Score and Suggestion. Assign a \texttt{confidence\_score} and suggest specific improvements.\\

**Details for Comparison:**\\
- **Previous Problem:**: \\\{last\_question\_details\} \\ 
- **Previous Expected Solution:**\\\{last\_question\_expected\_solution\}\\
- **New Problem:** \\\{new\_question\_details\} \\
- **New Expected Solution:**\\\{new\_question\_expected\_solution\}\\ 

**Result Format:**\\
Provide your evaluation in JSON format with these exact keys:  \\
\{\{\\
\quad "performance\_score": 0-100,\\
\quad "confidence\_score": 0-100\\
\}\}
\end{tcolorbox}

\begin{tcolorbox}[
    colframe = gray,
    colback = gray!5!white,
    coltitle = white,
    coltext = black,
    fonttitle = \bfseries,
    title = {Analyzing - level 4},
    boxrule = 1pt,
    arc = 2mm,
    width = \linewidth,
    left = 7pt,
    right = 7pt,
    top = 5pt,
    bottom = 5pt
]
\fontsize{8.5pt}{10pt}\selectfont
You are an expert in math and reasoning, acting as a refiner and evaluator, to assess the "Analyzing" level skills of a math problem generator by comparing a newly generated math problem with a previous one. \\

**Evaluation Criteria** \\
Please follow these steps: \\

Step 1: Identify "Big Five" Components: 1) math concepts and domains, 2) required skills to solve the problem, 3) math expressions as sequence of operations, 4) values used, and 5) creative narrative.\\
Step 2: Analyzing. Look for signs that the problem generator breaks down elements, highlights distinctions, and reorganizes structure.\\
Step 3: Levels of Analyzing:\\
- Strong Analyzing (80–100): Breaks down and reorganizes structure effectively to highlight deeper relationships.\\
- Medium Analyzing (60–80): Identifies structure but without major transformation.\\
- Low Analyzing (<60): Surface-level manipulation or copy with minimal analysis.\\
Step 4: Confidence Score and Suggestion. Assign a \texttt{confidence\_score} and suggest specific improvements.\\

**Details for Comparison:**\\
- **Previous Problem:**: \{last\_question\_details\} \\ 
- **Previous Expected Solution:**\{last\_question\_expected\_solution\}\\
- **New Problem:** \{new\_question\_details\} \\
- **New Expected Solution:**\{new\_question\_expected\_solution\}\\ 

**Result Format:**\\
Provide your evaluation in JSON format with these exact keys:  \\
\{\{\\
\quad "performance\_score": 0-100,\\
\quad "confidence\_score": 0-100\\
\}\}
\end{tcolorbox}

\begin{tcolorbox}[
    colframe = gray,
    colback = gray!5!white,
    coltitle = white,
    coltext = black,
    fonttitle = \bfseries,
    title = {Evaluating - level 5},
    boxrule = 1pt,
    arc = 2mm,
    width = \linewidth,
    left = 7pt,
    right = 7pt,
    top = 5pt,
    bottom = 5pt
]
\fontsize{8.5pt}{10pt}\selectfont
You are an expert in math and reasoning, acting as a refiner and evaluator, to assess the "Evaluating" level skills of a math problem generator by comparing a newly generated math problem with a previous one. \\

**Evaluation Criteria** \\
Please follow these steps: \\

Step 1: Identify "Big Five" Components: 1) math concepts and domains, 2) required skills to solve the problem, 3) math expressions as sequence of operations, 4) values used, and 5) creative narrative.\\
Step 2: Evaluating. Examine whether the generator makes justified choices, defends reasoning, and prioritizes design decisions.\\
Step 3: Levels of Evaluating:\\
- Strong Evaluating (80–100): Provides justified changes and demonstrates prioritization in design logic.\\
- Medium Evaluating (60–80): Modifies problem with some justifications or preference reasoning.\\
- Low Evaluating (<60): Minor edits without clear evaluation or rationale.\\
Step 4: Confidence Score and Suggestion. Assign a \texttt{confidence\_score} and suggest specific improvements.\\

**Details for Comparison:**\\
- **Previous Problem:**: \{last\_question\_details\} \\ 
- **Previous Expected Solution:**\{last\_question\_expected\_solution\}\\
- **New Problem:** \{new\_question\_details\} \\
- **New Expected Solution:**\{new\_question\_expected\_solution\}\\ 

**Result Format:**\\
Provide your evaluation in JSON format with these exact keys:  \\
\{\{\\
\quad "performance\_score": 0-100,\\
\quad "confidence\_score": 0-100\\
\}\}
\end{tcolorbox}

\begin{tcolorbox}[
    colframe = gray,
    colback = gray!5!white,
    coltitle = white,
    coltext = black,
    fonttitle = \bfseries,
    title = {Creating - level 6},
    boxrule = 1pt,
    arc = 2mm,
    width = \linewidth,
    left = 7pt,
    right = 7pt,
    top = 5pt,
    bottom = 5pt
]
\fontsize{8.5pt}{10pt}\selectfont
You are an expert in math and reasoning, acting as a refiner and evaluator, to assess the "Creating" level skills of a math problem generator by comparing a newly generated math problem with a previous one. \\

**Evaluation Criteria** \\
Please follow these steps: \\

Step 1: Identify "Big Five" Components: 1) math concepts and domains, 2) required skills to solve the problem, 3) math expressions as sequence of operations, 4) values used, and 5) creative narrative.\\
Step 2: Creating. Assess whether the generator develops original content by synthesizing and inventing meaningful structure or context.\\
Step 3: Levels of Creating:\\
- Strong Creating (80–100): Constructs novel and effective problem with well-integrated ideas.\\
- Medium Creating (60–80): Makes some changes or combinations with partial novelty.\\
- Low Creating (<60): Mostly rearranges or copies with minimal originality.\\
Step 4: Confidence Score and Suggestion. Assign a \texttt{confidence\_score} and suggest specific improvements.\\

**Details for Comparison:**\\
- **Previous Problem:**: \{last\_question\_details\} \\ 
- **Previous Expected Solution:**\{last\_question\_expected\_solution\}\\
- **New Problem:** \{new\_question\_details\} \\
- **New Expected Solution:**\{new\_question\_expected\_solution\}\\ 

**Result Format:**\\
Provide your evaluation in JSON format with these exact keys:  \\
\{\{\\
\quad "performance\_score": 0-100,\\
\quad "confidence\_score": 0-100\\
\}\}
\end{tcolorbox}

\subsection{Holistic Evaluation Agent - $A_7$}
\label{holistic agent prompt}
\begin{tcolorbox}[
    colframe = gray,
    colback = gray!5!white,
    coltitle = white,
    coltext = black,
    fonttitle = \bfseries,
    title = {Holistic Evaluation - General Quality},
    boxrule = 1pt,
    arc = 2mm,
    width = \linewidth,
    left = 7pt,
    right = 7pt,
    top = 5pt,
    bottom = 5pt
]
\fontsize{8.5pt}{10pt}\selectfont
You are an expert evaluator assessing Math Problem Quality and Math Language Quality in the educational question generation research context.\\ 
Please evaluate the quality of the following math word problem by analyzing its big five components and linguistic features. Identify and categorize any linguistic-level errors (e.g., ambiguity, unanswerability, or linguistic complexity) and assess the problem’s solution strategy. \\

**Details for Comparison:** \\
- **Previous Problem:** \{last\_question\_details\} \\
- **Previous Expected Solution:** \{last\_question\_expected\_solution\} \\
- **New Problem:** \{new\_question\_details\} \\
- **New Expected Solution:** \{new\_question\_expected\_solution\} \\

**Step 1: Big Five Components Extraction**\\
1) Math concepts and domains \\
2) Required skills to solve the problem \\
3) Math expressions as sequence of operations \\
4) Values that substitute into expressions \\
5) The narrative story based on real-life socio-cultural experiences \\

**Step 2: Lexical and Syntactic Complexity Analysis**\\
- Type-Token Ratio (TTR) \\
- Yngve Score \\
- Frazier Score \\
- Frazier–Roark Score \\
- Developmental Level \\
- Syntactic Frequency \\
- Mean Dependency Distance (MDD) \\
- Sentence Length \\

**Step 3: Error Identification and Classification**\\
- Ambiguity \\
- Unanswerability \\
- Rationality \\

**Step 4: Solution Strategy Analysis**\\
- One-Step or Multi-Step \\
- Comprehension Challenges from Multi-Step Reasoning \\

**Step 5: Improvement Suggestions**\\
Suggestions should address: \\
- Ambiguous phrasing \\
- Unanswerable problems \\
- Linguistic complexity \\
- Structure consistency and narrative realism \\

**Step 6: Performance Score Calculation (0–100)**\\
1. Lexical and Syntactic Complexity \\
2. Error Count and Severity \\
3. Clarity and Solvability \\
4. Answerability Penalty \\
5. Structural Consistency and Creativity \\

**Scoring Guidance:**\\
- 90–100: Clear, simple, and error-free problem. \\
- 70–89: Minor complexity or errors that slightly impact clarity. \\
- 50–69: Moderate complexity and multiple identifiable issues. \\
- 0–49: Significant errors, ambiguity, or unanswerable conditions. \\

**Result Format:**\\
Please return your evaluation in the following JSON format: \\
\{\{\\
\quad "performance\_score": 0-100,\\
\quad "confidence\_score": 0-100,\\
\quad "improvement\_suggestions": ["suggestion1", "suggestion2"]\\
\}\}
\end{tcolorbox}

\section{Human Expert Quality Evaluation}
\label{human}
\begin{table}[H]
\centering
\begin{minipage}{\linewidth}
\setlength{\tabcolsep}{4pt}
\renewcommand{\arraystretch}{1.15}
\footnotesize
\begin{tabular}{>{\raggedright\arraybackslash}p{0.18\linewidth} | >{\raggedright\arraybackslash}p{0.26\linewidth} | >{\raggedright\arraybackslash}p{0.26\linewidth} | >{\raggedright\arraybackslash}p{0.26\linewidth}}
\hline
\rowcolor[gray]{0.95} \textbf{Evaluation Component} & \textbf{Zero-shot Qwen2.5-14B} & \textbf{\textsc{THiNK}-Guided Qwen2.5-14B} & \textbf{Comparison \& Insight} \\ 
\hline
\textbf{Math Concepts and Domains} &
Implies inverse proportionality between number of musicians and performance time, akin to shared work problems in algebra. &
Recognizes invariance of musical performance duration, aligning with real-world temporal constraints rather than mathematical proportional reasoning. &
The baseline activates inappropriate algebraic domain reasoning, while the instruction-tuned version correctly disengages from it, reflecting conceptual coherence. \\ 
\hline
\rowcolor[gray]{0.98} \textbf{Prerequisite Skills} &
Requires procedural knowledge of ratio and unit manipulation but misapplies them due to the incorrect premise. &
Requires conceptual understanding of real-life constraints rather than computation. &
The instruction-tuned version activates domain-appropriate prior knowledge, indicating better alignment with relevant mental schemas. \\
\hline
\textbf{Mathematical Representations} &
Suggests (implicitly) a proportional formula: (120 musicians × 40 minutes) ÷ 60 = 80 minutes. No explicit expression, but logic implies computation. &
No symbolic expression: relies on verbal conceptual reasoning that performance time is independent of musician count if ensemble is complete. &
The baseline attempts structured reasoning but misapplies it; the tuned version avoids misleading formalism, showing better traceability and logic. \\ 
\hline
\rowcolor[gray]{0.98} \textbf{Alternative Values} &
Fails to generalize: if given different but equivalent values, the baseline would still apply faulty proportional logic. &
Generalizes correctly: the model recognizes that performance duration is invariant under alternative numbers of musicians, assuming parts are covered. &
Instruction tuning enhances generalization across input permutations that preserve the core problem structure. \\
\hline
\textbf{Narrative Stories} &
Uses a formal orchestra setting but leverages it in a way that misleadingly maps to mathematical workload sharing. &
Uses a school band narrative, maintaining realism while correctly situating the mathematical logic within a consistent real-world constraint. &
The instruction-tuned model better integrates narrative realism and reasoning integrity, supporting engagement without conceptual distortion. \\ 
\hline
\rowcolor[gray]{0.98} \textbf{Bloom's Taxonomy Level} &
\textit{Apply} (misapplied): Requires calculation, but the wrong concept leads to incorrect problem-solving. &
\textit{Understand / Analyze}: Requires unpacking implicit assumptions and applying invariant reasoning to a familiar context. &
The instruction-tuned version ascends Bloom's hierarchy, requiring abstract thinking and transfer, not mechanical execution. \\
\hline
\end{tabular}
\caption{Comparative analysis of baseline and instruction-tuned \textsc{Qwen2.5-14B-IT} models across multiple evaluation dimensions, highlighting improved contextual reasoning and domain-appropriate knowledge application.}
\label{human-eval}
\end{minipage}
\end{table}

\end{document}